# Harvesting comparable corpora and mining them for equivalent bilingual sentences using statistical classification and analogy-based heuristics

Krzysztof Wołk (✉), Emilia Rejmund, Krzysztof Marasek

Department of Multimedia
Polish - Japanese Academy of Information Technology
{kwolk, erejmund, kmarasek}@pja.edu.pl

**Abstract.** Parallel sentences are a relatively scarce but extremely useful resource for many applications including cross-lingual retrieval and statistical machine translation. This research explores our new methodologies for mining such data from previously obtained comparable corpora. The task is highly practical since non-parallel multilingual data exist in far greater quantities than parallel corpora, but parallel sentences are a much more useful resource. Here we propose a web crawling method for building subject-aligned comparable corpora from e.g. Wikipedia dumps and Euronews web page. The improvements in machine translation are shown on Polish-English language pair for various text domains. We also tested another method of building parallel corpora based on comparable corpora data. It lets automatically broad existing corpus of sentences from subject of corpora based on analogies between them.

## 1 Introduction

In this article we present methodologies that allow us to obtain truly parallel corpora from not sentence-aligned data sources, such as noisy-parallel or comparable corpora [1]. For this purpose we used a set of specialized tools for obtaining, aligning, extracting and filtering text data, combined together into a pipeline that allows us to complete the task. We present the results of our initial experiments based on text samples obtained from the Wikipedia dumps and the Euronews web page. We chose the Wikipedia as a source of the data because of a large number of documents that it provides (1,047,423 articles on PL Wiki and 4,524,017 on EN, at the time of writing this article). Furthermore, Wikipedia contains not only comparable documents, but also some documents that are translations of each other. The quality of our approach is measured by improvements in MT results.

Second method is based on sequential analogy detection. We seek to obtain parallel corpora from unaligned data. Solution proposed by our team is based on sequential analogy detection. Such approach was presented in literature [2][3], but all applications concern similar languages with similar grammar like English-French, Chinese-Japanese. We try to apply this method for English-Polish corpora. These two languages have different grammar, which makes our approach innovative and let easily broad this method for different languages pairs. In our approach, to enhance quality of identified analogies, sequential analogies clusters are sought.

## 2 State of the art

Two main approaches for building comparable corpora can be distinguished. Probably the most common approach is based on the retrieval of the cross-lingual information. In the second approach, source documents need to be translated using any machine translation system. The documents translated in that process are then compared with documents written in the target language in order to find the most similar document pairs.

The authors in [4] suggested obtaining only title and some meta-information, such as publication date and time for each document instead of its full contents in order to reduce the cost of building the comparable corpora (CC). The cosine similarity of titles term frequency vectors were used to match titles and contents of matched pairs.

An interesting idea for mining parallel data from Wikipedia was described in [5]. The authors propose in their word two separate approaches. The first idea is to use an online machine translation (MT) system to translate Dutch pages of the Wikipedia into English and they try to compare original EN pages with translated ones.

The authors of [6] facilitate a BootCat method that was proven to be fast and effective when a corpus building is concerned. The authors try to extend this method by adding support for multilingual data and also present a pivot evaluation.

Interwiki links were facilitated by Tyers and Pienaar in [7]. Based on the Wikipedia link structure a bilingual dictionary is extracted. In their work they measured that depending on the language pair the mismatch between linked Wikipedia pages averages.

What is more, authors of [8] introduce an automatic alignment method of parallel text fragments by using a textual entailment technique and a phrase-base Statistical Machine Translation (SMT) system. Authors state that significant improvement in SMT quality by using obtained data was obtained (increase in BLEU by 1.73).

## 3 Preparation of the data

Our procedure starts with a specialized web crawler implemented by us. Because PL Wiki contains less data of which almost all articles have their correspondence on EN Wiki, the program crawls data starting from non-English site first. The crawler can obtain and save bilingual articles of any language supported by the Wikipedia. The tool requires at least two Wikipedia dumps on different languages and information about language links between the articles in the dumps (obtained from the interwiki links). For the Euronews.com a standard web crawler was used. This web crawler was designed to use the Euronews.com archive page. In first phase it generates a database of parallel articles in two selected languages in order to collect comparable data from it.

For the experiments in the statistical machine translation we choose TED lectures domain, to be more specific the PL-EN TED[1] corpora prepared for IWSLT (International Workshop on Spoken Language Translation) 2014 evaluation campaign by the FBK (Fondazione Bruno Kessler). This domain is very wide and covers many not related subjects and areas. The data contains almost 2,5M untokenized words [9]. Additionally we choose two more narrow domains: The first parallel corpus is made out of PDF documents from the European Medicines Agency (EMEA) and medicine leaflets [10]. The second was extracted from the proceedings of the European Parliament (EUP)

---

[1] https://www.ted.com/talks

[11]. What is more we also conducted experiments on the Basic Travel Expression Corpus (BTEC), a multilingual speech corpus containing tourism-related sentences similar to those that are usually found in phrasebooks for tourists going abroad [12]. Lastly we used a corpus built from the movie subtitles (OPEN) [10].

In Table 1 we present details about number of unique words (WORDS) and their forms as well as about number of bilingual sentence pairs (PAIRS).

As mentioned, the solution can be divided into three main steps. First the data is collected, then it is aligned at article level, and lastly the results of the alignment are mined for parallel sentences. Sentence alignment must be computationally feasible in order to be of practical use in various applications [13].

| CORPORA | PL WORDS | EN WORDS | PAIRS |
| --- | --- | --- | --- |
| BTEC | 50,782 | 24,662 | 220,730 |
| TED | 218,426 | 104,117 | 151,288 |
| EMEA | 148,230 | 109,361 | 1,046,764 |
| EUP | 311,654 | 136,597 | 632,565 |
| OPEN | 1,236,088 | 749,300 | 33,570,553 |

*Table 1 Corpora specification*

With this methodology we were able to obtain 4,498 topic-aligned articles from the Euronews and 492,906 from the Wikipedia.

## 4 Parallel data mining

In order to extract the parallel sentence pairs we decided to try two different strategies. The first one facilitates and extends methods used in Yalign Tool[2] and the second is based on analogy detection. The MT results we present in this article were obtained with the first strategy. The second method is still in development phase, nevertheless the initial results are promising and worth mentioning.

### 4.1 Improvements to Yalign's method

In Yalign tool [14] for the sequence alignment A* search algorithm is used [15] to find an optimal alignment between the sentences in two given documents. Unfortunately it can't handle alignments that cross each other or alignments from two sentences into a single one [15]. To overcome this and other minor problems, in order to improve mining quality, we used the Needleman-Wunch algorithm (originally used for DNA sequences) instead. Because it would require N * M calls to the sentence similarity matrix we implemented its GPU version for accelerated processing [16].

The classifier must be trained in order to determine if a pair of sentences is translation of each other or not. The particular classifier used in this research was a Support Vector Machine [17].

What is more our solution facilitated multithreading and proved to increase the mining time by the factor of 5 in comparison with standard Yalign tool (using Core i7 CPU).

To train the classifier a good quality parallel data was necessary as well as a dictionary with translation probability included. For this purposes we used TED talks [18] corpora enhanced by us during the IWSLT'13 Evaluation Campaign [13]. In order to obtain a dictionary we trained a phrase table and extracted 1-grams from it. We used the MGIZA++ tool for word and phrase alignment. The lexical reordering was set to use the msd-bidirectional-fe method and the symmetrization method was set to grow-diag-final-and for word alignment processing [20]. For bi-lingual training data we used four corpora previously described. We obtained four different classifiers and repeated mining procedure with each of them. The detailed results for the Wiki are showed in Table 2.

---

[2] https://github.com/machinalis/yalign

| Classifier | Value | PL | EN |
| --- | --- | --- | --- |
| TED | Size in MB | 41,0 | 41,2 |
| | No. of sentences | 357,931 | 357,931 |
| | No. of words | 5,677,504 | 6,372,017 |
| | No. of unique words | 812,370 | 741,463 |
| BTEC | Size in MB | 3,2 | 3,2 |
| | No. of sentences | 41,737 | 41,737 |
| | No. of words | 439,550 | 473,084 |
| | No. of unique words | 139,454 | 127,820 |
| EMEA | Size in MB | 0,15 | 0,14 |
| | No. of sentences | 1,507 | 1,507 |
| | No. of words | 18,301 | 21,616 |
| | No. of unique words | 7,162 | 5,352 |
| EUP | Size in MB | 8,0 | 8,1 |
| | No. of sentences | 74,295 | 74,295 |
| | No. of words | 1,118,167 | 1,203,307 |
| | No. of unique words | 257,338 | 242,899 |
| OPEN | Size in MB | 5,8 | 5,7 |
| | No. of sentences | 25,704 | 25,704 |
| | No. of words | 779,420 | 854,106 |
| | No. of unique words | 219,965 | 198,599 |

*Table 2 Data mined from the Wikipedia for each classifier*

### 4.2 Analogy based method

This method is based on sequential analogy detection. Based on parallel corpus we detect analogies that exists in both languages. To enhance quality of identified analogies sequential analogies clusters are sought.

However our current research on Wikipedia corpora shows that it is both extremely difficult and machine time consuming to seek out clusters of higher orders. Therefore we restrained ourselves to simple analogies such as A is to B in the same way as C to D.

$$A:B::C:D$$

Such analogies are found using distance calculation. We seek such sentences that:

$$dist(A,B)=dist(C,D)$$

and

$$dist(A,C)=dist(B,D)$$

Additional constrain was added that requires the same relation of occurrences of each character in the sentences. E.g. if number of character "a" in sentence A is equal to x and equal to y in sentence B then the same relation must occur in sentences C and D.

We used Levenshtein metric in our distance calculation. We tried to apply it directly into the characters in the sentence, or considering each word in the sentence, as individual symbol, and calculating Levenshtein distance between symbol coded sentences. The latter approach was employed due to the fact that this method was earlier tested on Chinese and Japanese languages [19] that use symbols to represent entire words.

After clustering, data from clusters are compared to each other to find similarities between them. For each four sentences

$$A:B::C:D$$

We look for such E and F that:

$$C:D::E:F \text{ and } E:F::A:B$$

However none were found in our corpus, therefore we restrained ourselves to small clusters with size of 2 pairs of sentences. In every cluster matching sentences from parallel corpus were identified. It let us generate new sentences similar to the one which are in our corpus and add it to broad resulting data set. For each of sequential analogies that were identified, rewriting model is constructed. This is achieved by string manipulation. Common pre- and suffixes for each of the sentences are calculated using LCS (Longest Common Subsequence) method.

Sample of rewriting model is shown on this example (prefix and suffix are shown in bold)

*__Poproszę__ koc i poduszkę.* ⇔ *__A__ blanket and a pillow__, please.__*

*__Czy mogę poprosić__ o śmietankę i cukier?* ⇔ *__Can I have__ cream and sugar?*

Rewriting model consist of prefix, suffix and their translation. It is now possible to construct parallel corpus form non-parallel monolingual source. Each sentence in the corpus is tested for match with the model. If the sentence contains prefix and suffix is considered matching sentence.

*__Poproszę__ bilet.* ⇔ *__A__ unknown__, please.__*

In the matched sentence some of the words remain not translated but general meaning of the sentence is conveyed. Remaining words may be translated word-by-word while translated sentence would remain grammatically correct.

bilet ⇔ ticket

Substituting unknown words with translated ones we are able to create a parallel corpus entry.

*__Poproszę__ bilet.* ⇔ *__A__ ticket__, please.__*

As a result of sequential analogy detection based method we mined **8128** models from of Wikipedia parallel corpus. This enabled generation of **114,000** new pair sentences to extend parallel corpus. Sentences were generated from Wikipedia comparable corpus that is basically an extract of Wikipedia articles. Therefore we have article in Polish and English on the same topic, but sentences are not aligned in any particular way. We use rewriting models to match sentences from Polish article to sentences in English. Whenever model could be successfully applied to a pair of sentences, this pair is considered to be parallel resulting in generation of quasi-parallel corpus (quasi since sentences were aligned artificially using approach described above). Those parallel sentences can be used to extend parallel corpora in order to improve quality of translation.

## 5 Results and evaluation

In order to evaluate the corpora we divided each corpus into 200 segments and randomly selected 10 sentences from each segment for testing purposes. This methodology ensured that the test sets covered entire corpus. The selected sentences were removed from the corpora. We trained the baseline system, as well as system with extended training data with the Wikipedia corpora and lastly we used Modified Moore Levis Filtering for the Wikipedia corpora domain adaptation. Additionally we used monolingual part of the corpora as language model and we tried to adapt it for each corpus by using linear interpolation [2]. For scoring purposes we used four well-known metrics that show high correlation with human judgments. Among the commonly used SMT metrics are: Bilingual Evaluation Understudy (BLEU) [11] the U.S. National Institute of Standards & Technology (NIST) metric [20], the Metric for Evaluation of Translation with Explicit Ordering (METEOR) [8], and Translation Error Rate (TER) [20].

The baseline system testing was done using the Moses open source SMT toolkit with its Experiment Management System (EMS) [16] with settings described in [13].

Starting from baseline systems (BASE) tests in PL to EN and EN to PL direction, we raised our score through extending the language model (LM), interpolating it (ILM) and by the corpora extension with additional data (EXT) and by filtering additional data with Modified Moore Levis Filtering (MML) [2]. It must be noted that extension of language models was done on systems with corpora after MML filtration. The LM and ILM experiments already contain extended training data. The results of the experiments are showed in Table 3.

|        |        | Polish to English | | | | English to Polish | | | |
|--------|--------|-------|-------|-------|--------|-------|-------|-------|--------|
| Corpus | System | BLEU  | NIST  | TER   | METEOR | BLEU  | NIST  | TER   | METEOR |
| TED    | BASE   | 16,96 | 5,26  | 67,10 | 49,42  | 10,99 | 3,95  | 74,87 | 33,64  |
|        | EXT    | 16,96 | 5,29  | 66,53 | 49,66  | 10,86 | 3,84  | 75,67 | 33,80  |
|        | MML    | 16,84 | 5,25  | 67,55 | 49,31  | 11,01 | 3,97  | 74,12 | 33,77  |
|        | LM     | 17,14 | 5,27  | 67,66 | 49,95  | 11,54 | 4,01  | 73,93 | 34,12  |
|        | **ILM**| **17,64** | **5,48** | **64,35** | **51,19** | **11,86** | **4,14** | **73,12** | **34,23** |
| BTEC   | BASE   | 11,20 | 3,38  | 77,35 | 33,20  | 8,66  | 2,73  | 85,27 | 27,22  |
|        | EXT    | 12,96 | 3,72  | 74,58 | 38,69  | 8,46  | 2,71  | 84,45 | 27,14  |
|        | MML    | 12,80 | 3,71  | 76,12 | 38,40  | 8,50  | 2,74  | 83,84 | 27,30  |
|        | LM     | 13,23 | 3,78  | 75,68 | 39,16  | 8,76  | 2,78  | 82,30 | 27,39  |
|        | **ILM**| **13,60** | **3,88** | **74,96** | **39,94** | **9,13** | **2,86** | **82,65** | **28,29** |
| EMEA   | BASE   | 62,60 | 10,19 | 36,06 | 77,48  | **56,39** | **9,41** | **40,88** | **70,38** |
|        | EXT    | 62,41 | 10,18 | 36,15 | 77,27  | 55,61 | 9,28  | 42,15 | 69,47  |
|        | MML    | 62,72 | 10,24 | 35,98 | 77,47  | 55,52 | 9,26  | 42,18 | 69,23  |
|        | LM     | 62,90 | 10,24 | 35,73 | 77,63  | 55,38 | 9,23  | 42,58 | 69,10  |
|        | **ILM**| **62,93** | **10,27** | **35,48** | **77,87** | 55,62 | 9,30  | 42,05 | 69,61  |
| EUP    | **BASE**| **36,73** | **8,38** | **47,10** | **70,94** | **25,74** | **6,54** | **58,08** | **48,46** |
|        | EXT    | 36,16 | 8,24  | 47,89 | 70,37  | 24,93 | 6,38  | 59,40 | 47,44  |
|        | MML    | 36,66 | 8,32  | 47,25 | 70,65  | 24,88 | 6,38  | 59,34 | 47,40  |
|        | LM     | 36,69 | 8,34  | 47,13 | 70,67  | 24,64 | 6,33  | 59,74 | 47,24  |
|        | ILM    | 36,72 | 8,34  | 47,28 | 70,79  | 24,94 | 6,41  | 59,27 | 47,64  |
| OPEN   | BASE   | 64,54 | 9,61  | 32,38 | 77,29  | **31,55** | **5,46** | **62,24** | **47,47** |
|        | EXT    | 65,49 | 9,73  | 32,49 | 77,27  | 31,49 | 5,46  | 62,06 | 47,26  |
|        | MML    | 65,16 | 9,62  | 33,79 | 76,45  | 31,33 | 5,46  | 62,13 | 47,31  |
|        | LM     | 65,53 | 9,70  | 32,94 | 77,00  | 31,22 | 5,46  | 62,61 | 47,29  |
|        | **ILM**| **65,87** | **9,74** | **32,89** | **77,08** | 31,39 | 5,46  | 62,43 | 47,33  |

*Table 3 Polish to English and English to Polish MT Experiments*

## 6 Conclusions

The results showed in Table 4 and Table 5, to be more specific BLEU, Meteor and TER values in TED corpus were checked whether the differences were relevant. We measured the variance due to the BASE and MML set selection. It was calculated using bootstrap resampling[3] for each test run. The result for BLEU was 0.5 and 0.3 and 0.6 for METEOR and TER respectively. The results over 0 mean that there is significant (to some extent) difference between the test sets and it indicates that a difference of this magnitude is likely to be generated again by some random translation process, which would most likely lead to better translation results in general. [21]

The results of SMT systems based only on mined data were not too surprising. Firstly, they confirm quality and high parallelism level of the corpora that can be concluded from the translation quality especially on the TED data set. Only 2 BLEU points gap can be observed when comparing systems trained on strict in-domain (TED) data and mined data, when it comes to EN – PL translation system. It also seems natural that the best SMT scores were obtained on TED data. It is not only most similar to the Wikipedia articles and overlaps with it in many topics but also the classifier trained on the TED data set recognized most of parallel sentences. In the results it can also be observed that the METEOR metric in some cases rises whereas other metrics decrease. Most likely reason for this is fact that other metrics suffer, in comparison to the METEOR, from the lack of scoring mechanism for synonyms. The Wikipedia is very wide not only when we consider its topics but also vocabulary, which leads to conclusions that mined corpora, is good source for extending sparse text domains. It is also the reason why the test sets originating from wide domains outscore narrow domain ones and also the most likely explanation why training on larger mined data sometimes slightly decreases test sets from very specific domains. Nonetheless it must be noted that after manual analysis we conceded that in many cases translations were good but automatic metric became lower because of the usage of synonyms.

Nowadays, the bi-sentence extraction task is becoming more and more popular in unsupervised learning for numerous specific tasks. The method overcomes disparities between two languages. It is a language independent method that can easily be adjusted to a new environment, and it only requires parallel corpora for initial training. The experiments show that the method performs well. The obtained corpora increased MT quality in wide text domains. From a practical point of view, the method neither requires expensive training nor requires language-specific grammatical resources, while producing satisfying results.

---

[3] https://github.com/jhclark/multeval